\pdfoutput=1

\documentclass[11pt]{article}

\usepackage[final]{EMNLP2023}

\usepackage{times}
\usepackage{latexsym}

\usepackage[T1]{fontenc}

\usepackage[utf8]{inputenc}
\usepackage{multirow}
\usepackage{microtype}
\usepackage{multicol}
\usepackage{inconsolata}
\usepackage{booktabs}
\usepackage{caption}
\usepackage{makecell}
\usepackage{tabularx}
\usepackage{multirow}

\usepackage{xcolor}
\usepackage[linesnumbered,ruled,vlined]{algorithm2e}

\usepackage{longtable}

\usepackage{graphics}
\usepackage{graphicx}

\usepackage{stfloats}

\title{KICGPT: Large Language Model with Knowledge in Context for Knowledge Graph Completion }

\author{Yanbin Wei \textsuperscript{1, 2}, Qiushi Huang\textsuperscript{1,3}, Yu Zhang\textsuperscript{1}\thanks{\ \ Corresponding author}\ , James T. Kwok\textsuperscript{2} \\
\textsuperscript{1} Southern University of Science and Technology \\
  \textsuperscript{2} Hong Kong University of Science and Technology \\
  \textsuperscript{3} University of Surrey}

\begin{document}
\maketitle
\begin{abstract}

Knowledge Graph Completion (KGC) is crucial for addressing knowledge graph
incompleteness and supporting downstream applications. Many models have been
proposed for KGC. They can be categorized into two main classes: triple-based and
text-based approaches. Triple-based methods struggle with long-tail entities due
to limited structural information and imbalanced entity distributions. Text-based
methods alleviate this issue but require costly training for language models and
specific finetuning for knowledge graphs, which limits their efficiency. To
alleviate these limitations, in this paper, we propose \textbf{KICGPT}, a
framework that integrates a large language model (LLM) and a triple-based KGC
retriever. It alleviates the long-tail problem without incurring additional
training overhead. KICGPT uses an in-context learning strategy called
\textbf{Knowledge Prompt}, which encodes structural knowledge into demonstrations
to guide the LLM. Empirical results on benchmark datasets demonstrate the
effectiveness of KICGPT with smaller training overhead and no finetuning. Code and data will be available at https://github.com/WEIYanbin1999/KICGPT.

\end{abstract}

\section{Introduction}

Knowledge Graphs (KGs) are powerful representations of real-world knowledge. 
The relationships among entities
are captured
by triples  of the form <head entity, relation, tail entity>.
KGs serve as the foundation for various applications such as recommendation systems, question-answering, and knowledge discovery. Knowledge Graph Completion (KGC) plays a crucial role for KGs by completing incomplete triples and hence addressing the inherent incompleteness of KGs. This paper focuses on the link prediction task in KGC, which is to predict
the missing entity in an incomplete triple.

Based on the source of information used, existing KGC methods can be categorized
into two main classes: triple-based methods and text-based methods \citep{wang2022language}. Triple-based methods (e.g., TransE \citep{bordes2013translating}, R-GCN \citep{schlichtkrull2018modeling}, and HittER \citep{chen-etal-2021-hitter}) utilize the structure of the knowledge graph as the only source of information for KGC. 
Given the typical imbalance of entities in KGs, long-tail entities are prevalent
and the KGC task has limited structural information about them.
Consequently, due to information scarcity, the performance of triple-based methods tends to degrade when processing long-tail entities \cite{wang2022language}. 
Text-based methods (such as KG-BERT \citep{yao2019kg}) mitigate this information
scarcity problem by encoding textual description as an extra information source \citep{wang2022language}. 
Besides the use of pre-trained language models (PLMs), finetuning remains
necessary for text-based methods to handle diverse knowledge graphs. However, this
is resource-intensive and requires task-specific finetuning for each downstream task.

Large language models (LLMs), such as ChatGPT and GPT-4 \citep{openai2023gpt}, have
extensive internal knowledge repositories from their vast pre-training corpora, which
can be used as an extra knowledge base to alleviate information scarcity for
long-tail entities. 
\citeauthor{tay2022transcending}
(\citeyear{tay2022transcending}),
\citeauthor{ouyang2022training}
(\citeyear{ouyang2022training})
demonstrated the potential for LLMs to handle a wide range of tasks by
specially-designed prompts without requiring training or finetuning.  Given the appealing properties
of LLMs, a straightforward approach is to directly apply LLMs to KGC tasks.
However, empirical evidence \citep{zhu2023llms} indicates that pure LLMs still
cannot achieve state-of-the-art performance in KGC tasks such as link prediction.
Additionally, there are several challenges that hinder the application of
LLM on KGC tasks. 
First, the LLM outputs can be unconstrained and may fall outside the scope of
entities in the KGs. Second,
LLMs impose length limits on the input tokens, and the limits are far from sufficient for describing a complete KGC task. Lastly, there is no effective in-context learning prompt design for LLM on KGC tasks.  

To alleviate the aforementioned limitations,  
we propose 
in this paper
a \textbf{K}nowledge \textbf{I}n \textbf{C}ontext with \textbf{GPT}
(\textbf{KICGPT}) framework. This integrates LLMs with a traditional
structure-aware KG model (which is called a retriever). Specifically, for each
query $q = (h,r,?)$ or $q = (?,r,t)$ in the link prediction task, where ``$?$"
denotes the missing tail or head entity to be predicted, the retriever first
processes the query $q$ independently and generates an ordered candidate entity
list $R_{retriever}$ that ranks all entities in the KG based on their retrieval
scores. The LLM then performs re-ranking on the top $m$ entities returned by
$R_{retriever}$, and replaces these $m$ entities with re-ranked ones returned by the LLM as the final result $R_{KICGPT}$. 
To achieve the LLM re-ranking, we propose \textbf{Knowledge Prompt}, a strategy of
in-context learning (ICL) \citep{wei2022emergent}, to encode the KG knowledge into
demonstrations in prompts.
Note that in KICGPT we only need to train the retriever. 
Compared with existing triple-based methods, KICGPT enables simultaneous
utilization of knowledge sources (i.e., KG and LLM's knowledge base) and
facilitates their alignment and enrichment to alleviate information scarcity for long-tail entities. Moreover, different from existing text-based models, KICGPT leverages a much larger semantic knowledge base without incurring additional training overhead. 
Unlike directly applying the LLM to the KGC task, which may generate undesired
outputs, the proposed KICGPT constrains its output by formalizing link prediction as a re-ranking task for the given sequence $R_{retriever}$.
Moreover, the standard link prediction task requires ranking for all the entities,
which is not feasible for LLM due to the length limit on input tokens. To overcome
this limitation, KICGPT utilizes the retriever to obtain the top-$m$ entities in
$R_{retriever}$, and only allows the LLM to perform re-ranking on these entities.


In summary, our contributions are as follows.
\begin{itemize}
\item We propose a novel cost-effective
framework \textbf{KICGPT} for
KGC tasks. To the best of our knowledge, this is the first work
that combines LLMs with triple-based KGC methods, offering a unique solution to address the task.

\item We propose a novel in-context learning strategy, \textbf{Knowledge Prompt}, specifically designed for KGC. 

\item 
Extensive experiments on benchmark datasets demonstrate that KICGPT 
achieves state-of-the-art performance with low training overhead.
\end{itemize}

\section{Related Work}
\paragraph{Triple-based KGC} Most existing KGC methods are triple-based methods,
which complete the knowledge graph solely based on the triple information. Early
shallow knowledge graph embedding (KGE) methods represent entities and
relationships as low-dimensional embedding vectors in a continuous embedding
space. 
Based on the scoring function,
these methods can be further categorized 
\citep{wang2017knowledge}
as translation-based (e.g., TransE \citep{bordes2013translating}) and semantic matching models (e.g., RESCAL \cite{nickel2011three} and DistMult \cite{yang2014embedding}).

However, they suffer from limited expressive power due to the use of shallow network
structures.
In recent years, more powerful network structures are integrated to solve KGC tasks.
Examples include the Graph Neural Networks \citep{schlichtkrull2018modeling},
Convolutional Neural Networks \citep{dettmers2018convolutional}, and Transformer
\citep{chen-etal-2021-hitter}. Most of these aggregate local structure context
into node embeddings and achieve much improved performance. However, they are
still limited by the imbalanced distribution of knowledge graph structure with
insufficient knowledge about the long-tail entities. Meta-learning
\citep{xiong2018one, chen2019meta} and logical rules \citep{sadeghian2019drum} can
mitigate the long-tail problem in KG. The main difference between them and our
work is that they handle long-tail entities by extracting and summarizing common
structural patterns or rules from the limited information in KG, while KICGPT
combines a vast external knowledge base inside the LLM with the structural
information in KGs, which can help alleviate information scarcity. 

\paragraph{Text-based KGC}
In light of the success of Natural Language Processing (NLP), text-based knowledge
graph completion is gaining more attention. As another mode of knowledge different
from the structured KG, text can provide rich semantic information. DKRL
\citep{xie2016representation} first introduces textual descriptions into entity
embeddings produced by a convolutional neural network. Subsequent works (such as
KG-BERT \citep{yao2019kg}, KEPLER \citep{wang2021kepler}, and Pretrain-KGE
\citep{zhang2020pretrain}) use a pre-trained language model (PLM) to
encode the text descriptions. More recently, LMKE \citep{wang2022language}
proposed a contrastive learning framework that uses the PLM to obtain 
entity and relation
embeddings 
in the same space as word tokens, and demonstrated its effectiveness on long-tail
problem
These methods generally rely on the language models to process text descriptions and require finetuning for different knowledge graphs. The proposed KICGPT, which uses LLM directly, is more efficient because it is training-free and requires no finetuning.

\paragraph{LLMs for KGs} Some recent works also explore the ability of LLM on KG
tasks. StructGPT \citep{jiang2023structgpt} proposed a general framework for
improving zero-shot reasoning ability of LLMs on structured data. It leverages the
LLM to perform reasoning on the KG Question Answering (KGQA) task with the help of
auxiliary interfaces that fetch the needed information from KG. Though both
StructGPT and our work utilize KG and LLM, StructGPT aims to help the LLM to
handle structured data and explore KGQA tasks, while ours utilizes the knowledge
base inside the LLM to handle 
KGC's
long-tail problem. Moreover, StructGPT performs
multi-step reasoning directly based on the KG structure, while the proposed KICGPT
utilizes the KG information in a different way. First, we use the whole KG to
generate preliminary results before using the LLM. Moreover, we incorporate a
portion of the KG triples into ICL demonstrations to help the LLM conduct reasoning. 
\citeauthor{zhu2023llms} (\citeyear{zhu2023llms}) directly evaluated the
performance of a LLM on KG reasoning.
Our work design an ICL strategy to guide the LLM to perform reasoning. Moreover,
their experiments on the link prediction task only involve 25 sampled instances
from the FB15k-237 dataset (which has 20,466 test triples).
Instead, we re-run their setting on ChatGPT and report the new results as a baseline in our experiments.

\paragraph{ICL for LLMs}
A typical way of approaching LLM is through in-context learning \cite{gpt3}, by
providing explicit instructions and demonstrations to guide the model's behavior.
This approach has been effective in various language understanding and generation tasks \cite{gopher, wei2022emergent}. In-context learning exposes the model to specific contextual information, allowing it to grasp and reproduce necessary patterns and structures for precise generation  \cite{ouyang2022training}. However, the success of in-context learning depends heavily on the quality of the prompt, and crafting suitable prompts can be delicate \cite{wang2022rationale}. While there has been work exploring ICL on different tasks \citep{dong2022survey}, to the best of our knowledge, no such work has been done for KGC. Unlike existing works, the proposed ICL strategy, Knowledge Prompt, considers the characteristics of KGC tasks and demonstrates its effectiveness on KGC tasks.

\section{Methodology}
In this section, we introduce the proposed KICGPT model. The complete algorithm is in appendix \ref{app: alg}.

\begin{figure*}[!ht]
    \includegraphics[width=\textwidth]{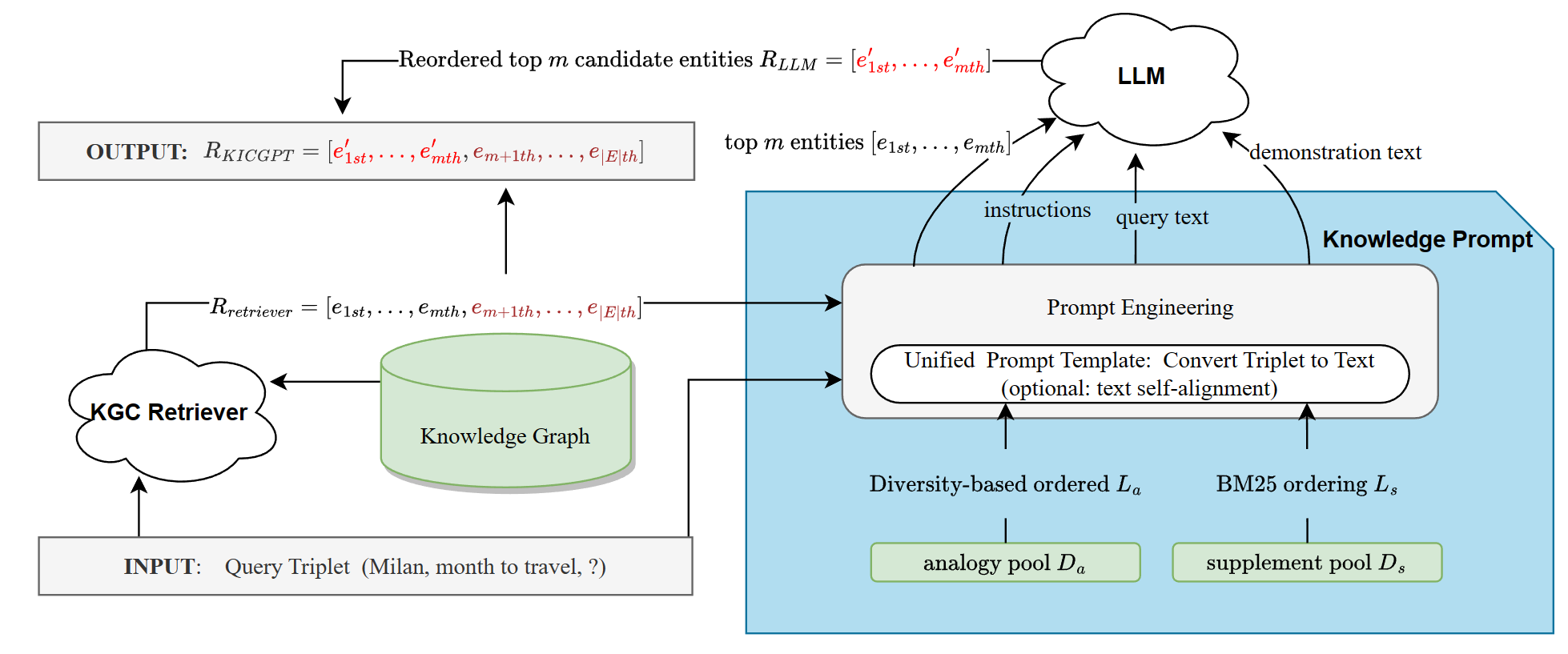}
    \caption{An illustration of the KICGPT framework.}
    \label{fig:1}
\end{figure*}

\subsection{Problem Setting} 
A knowledge graph can be represented as a set of triples $G = \{(h,r,t)\}$, where
$E$ and $R$ denote the set of entities and relations in $G$, respectively, $h \in
E$ is a head entity, $t \in E$ is a tail entity, and $r \in R$ represents the
relation between them. Link prediction is an important task in KGC. Given an
incomplete triple $(h,r,?)$ or $(?, r,t)$ as query, link prediction aims to
predict the missing entity (denoted as $?$). A link prediction model usually needs
to score all plausible entities as missing entity and then rank all entities in
descending order. For simplicity of presentation, we focus on queries with missing
tail entities (i.e., $(h,r,?)$). Queries with missing head entities (i.e.,
$(?,r,t)$) can be handled analogously.

\subsection{Overview}

Figure \ref{fig:1} illustrates the KICGPT framework.
There are two components: a triple-based KGC retriever and a LLM. 
For each query triple $(h,r,?)$, the retriever first generates the score of
$(h,r,e)$ for each entity $e \in E$. The ranking (in descending score) 
of all the entities 
is denoted $R_{retriever} = [e_{1st},e_{2nd},e_{3rd},\dots, e_{|E|th}]$. 
The LLM then performs a re-ranking of the 
top-$m$ entities 
based on its knowledge and demonstrations in the proposed Knowledge Prompt.
The re-ranking 
$R_{LLM} = [e'_{1st},e'_{2nd},e'_{3rd},\dots,e'_{mth}]$ is a permutation of
$[e_{1st},e_{2nd},e_{3rd},\dots,e_{mth}]$. By replacing the $m$ leading entities
in $R_{retriever}$ by $R_{LLM}$, the KICGPT outputs $R_{KICGPT} = [e'_{1st},e'_{2nd},e'_{3rd},\dots,e'_{mth},e_{m+1th},\dots,e_{|E|th}]$.

\subsection{Knowledge Prompt}
\label{sec: knowledge prompt}

In this section, we introduce 
Knowledge Prompt, 
an in-context learning strategy 
specially designed for KGC tasks. By encoding part of
the KG into demonstrations, Knowledge Prompt boosts the performance of LLM on link
prediction. 

\subsubsection{Demonstration Pool}
\label{pool}
For each query $(h,r,?)$,
we construct two pools of triples, $D_{a}$ and $D_{s}$, from the KG as demonstrations.

\paragraph{Analogy Pool} The analogy pool $D_{a}$ contains triples that help the
LLM to better understand the semantics of the query by analogy. Let $G_{train}$
and $G_{valid}$ be the sets of KG triples for training and validation,
respectively. $D_{a} = \{ (e',r,e'') \in G_{train} \cup G_{valid}| e',e''\in E\}$
includes triples with the same relation as the query $(h,r,?)$. 

\paragraph{Supplement Pool} The supplement pool $D_{s}$ contains triples which
provide supplementary information about the 
query's
head entity $h$.
Specifically, $D_{s}$
includes all triples with $h$ as the head or tail entity in the training and validation parts:
$\{(h,r',e') \in G_{train} \cup G_{valid}|r'\in R, e'\in E\} \cup \{(e',r',h) \in
G_{train} \cup G_{valid}|r'\in R, e'\in E\}$.

\subsubsection{Demonstration Ordering}
\label{demon order}

As the demonstration order affects performance \cite{ye2023compositional}, 
we propose different ordering strategies for the analogy and supplement pools.

For the analogy pool $D_{a}$, all its triples are similar to the
query because they share the same relation. As diversity of demonstrations is
important so that the LLM can learn from various analogy demonstrations, we propose an
ordering strategy that promotes diversity. Specifically, first we set a zero
counter for each entity. 
A triple from $D_{a}$ 
is randomly selected 
as 
demonstration, and the counters associated with its entities are increased by 1.
We iteratively choose 
as demonstration
the triple 
whose
entities' associated
counters
have the smallest sum
(tie is resolved 
randomly).
The associated counters 
are then increased
by 1. This is
repeated until all triples in $D_{a}$ are used, and the resultant
demonstration list obtained is denoted $L_{a}$.

For the supplement pool $D_{s}$, as it serves to provide supplementary
information on the 
query's
head entity $h$, we prefer related demonstrations
to the query. Specifically, we rank all triples in $D_{s}$ according to their BM25
scores \citep{robertson1994some}. This score is used to evaluate the correlation
between texts in each demonstration and query. The resultant ranking list, denoted $L_{s}$, contains all triples in $D_{s}$. 

Queries with a specific relation or head entity use a shared analogy or supplement pool. To prevent double counting, demonstration pools are created and ordered in data pre-processing. During inference, the corresponding pool is used based on the query.

\subsubsection{Prompt Engineering}
\label{prompt}
Prompt engineering is an important in ICL \citep{zhao2021calibrate}. In this
section, we show the whole interaction workflow with the LLM, and also
considerations of the prompt design. 

\begin{figure}[!ht]
    \includegraphics[width=\linewidth]{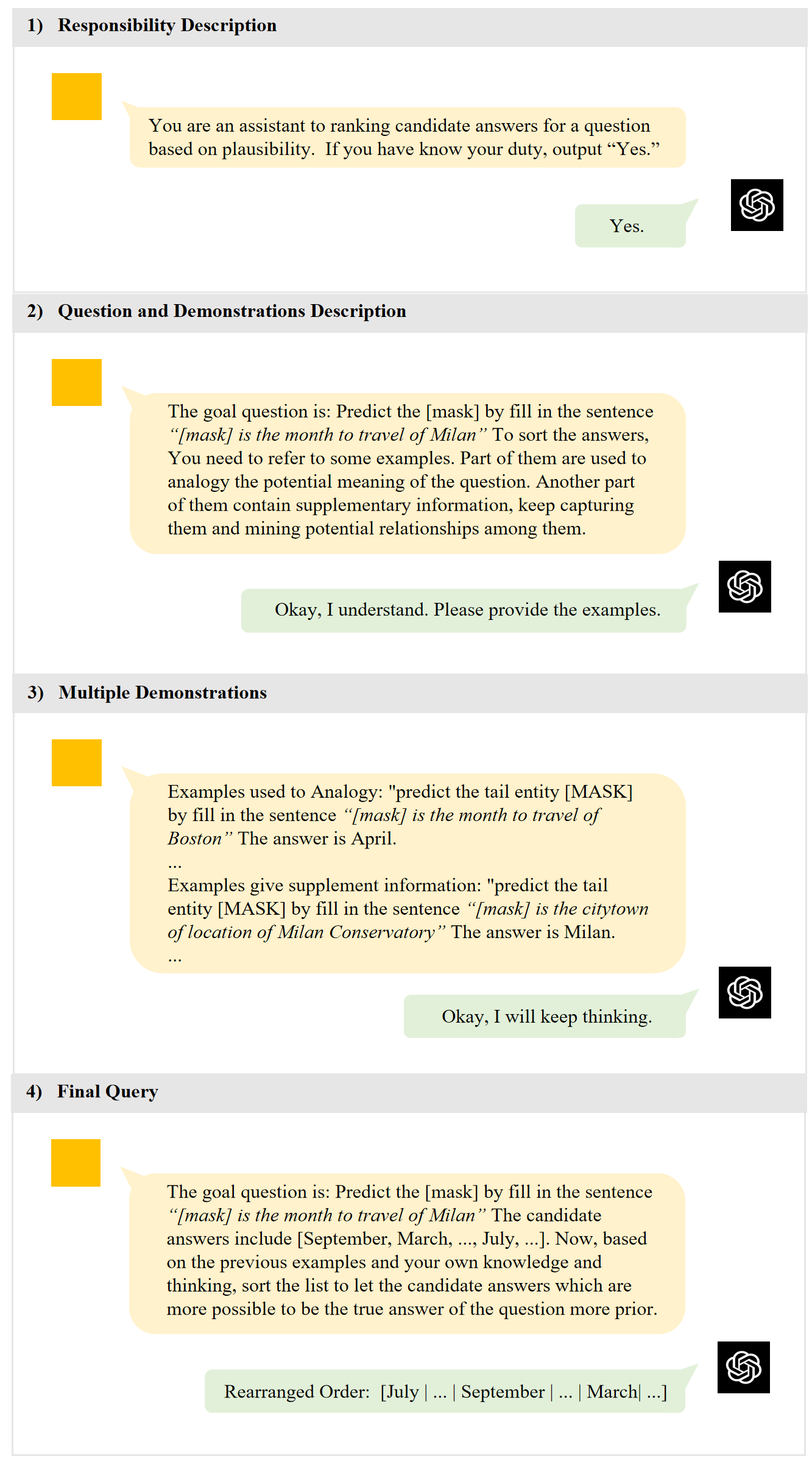}
    \caption{Illustration of a multi-round interaction with the LLM. Stage 3 is
	 repeated many times to provide more demonstrations.}
    \label{fig:flow}
\end{figure}

The demonstrations and query take the form of triples. However, LLMs require natural language input. To remedy this gap, KICGPT uses a unified prompt template to convert the query and demonstrations to plain text with the same format. 
We then perform multi-round interactions with the LLM, so as to guide it to
perform re-ranking, where these texts and some instructions are organized as
prompt inputs. Figure \ref{fig:flow} illustrates the workflow of a multi-round
interaction with the LLM. 
For each link prediction query $(h,r,?)$,
KICGPT creates an independent conversational prompt.
The whole multi-round interaction process includes four stages: responsibility description, question and demonstration description, multiple demonstrations, and final query. 

Responsibility description is illustrated in the
first part of Figure \ref{fig:flow}. We tell the LLM 
that its role is an assistant to ranking candidate answers for a question based on plausibility. 
We then check its feedback to ensure that it knows the task.

The question and demonstration description stage is shown in the second part of
Figure \ref{fig:flow}. We input the question (corresponding to the query's text)
and tell it that two types of examples are to be provided
and should be treated differently,
one for analogy and one
containing supplementary information.

Next,
illustrated in the third part of Figure \ref{fig:flow}, 
we provide the LLM with a batch of demonstrations ($L_{a}$ and $L_{s}$)
from the analogy pool and supplement pool.
To include more KG information and demonstrations, we repeat this
step as many times as possible subject to 
the input token length
limit.

Finally, we restate the query text, 
and ask
the LLM to re-rank 
the top-$m$ candidate entities.
The feedback is parsed into an ordered list $R_{LLM}$, which then replaces the
top-$m$ entities in $R_{retriever}$ as the final answer.

\subsection{Text Self-Alignment}
\label{text align}
In this section, we propose Text Self-alignment for KG text cleaning. It transforms the raw text in the KG to more understandable descriptions by the LLM.

Raw and obscure text descriptions generally exist in the KG. For example, a raw
relation description in the FB15k-237 dataset is
\textit{``/tv/tv\_program/country\_of\_origin''}. Most existing
methods \citep{yao2019kg,zhang2020pretrain,wang2021kepler,wang2022language}
convert it to a cleaner description by removing the symbols (e.g., \textit{``tv tv
program country of origin''}). By default, KICGPT uses ``of'' to organize such
hierarchical relation, leading to \textit{``country of origin of tv program of
tv''}. However, this description may still be
hard 
for the LLM 
to comprehend,
and may lead to incorrect responses.  

To address the above issue, 
we use 
ICL and let
the LLM to summarize natural text descriptions from the given
demonstrations that use raw descriptions. Specifically, for a relation $r$, we use
the analogy pool $D_{a}$ for $r$ (Section \ref{pool}) as demonstrations and order
them to get the ordered list $L_{a}$ (Section \ref{demon order}). These
demonstrations are fed as a prompt to the LLM
(Appendix \ref{sec:appendix}), 
which is asked to summarize the semantics of the relation into a sentence 
(the self-aligned text description of $r$). 
For the above example with raw relation text \textit{``/tv/tv\_program/country\_of\_origin''}, 
the LLM generates the 
much clearer and natural
self-aligned text description \textit{``$[T]$ is the country
where the TV program $[H]$ originated from.''},
where $[T]$ and $[H]$ are placeholders for the tail and head entities,
respectively.

Data pre-processing uses text self-alignment to create clear and aligned descriptions for each relation, which can be directly used as relation text in KICGPT for link prediction. This variant is called KICGPT$_{tsa}$. Since the texts are derived from the LLM, they conform to its presentation conventions, making them easier to understand and improving performance.
\section{Experiment}
\begin{table*}[!ht]
\center
\resizebox{\linewidth}{!}{
\begin{tabular}{|c|c|c|c|c|c|c|c|c|}\hline

\textbf{Dataset} & \multicolumn{4}{c|}{\textbf{FB15k-237}} & \multicolumn{4}{c|}{\textbf{WN18RR}} \\ \cline{1-9}
\textbf{Metric} & \textbf{MRR} & \textbf{Hits@1} & \textbf{Hits@3} & \textbf{Hits@10} & \textbf{MRR} & \textbf{Hits@1} & \textbf{Hits@3} & \textbf{Hits@10}\\ \cline{1-9} \hline

\multicolumn{9}{|c|}{\textbf{Triple-based methods}}\\ \hline

RESCAL\citep{nickel2011three} $\clubsuit$ & 0.356 & 0.266 & 0.390 & 0.535 & 0.467 & 0.439 & 0.478 & 0.516\\ \hline
TransE \citep{bordes2013translating} $\spadesuit$ & 0.279 & 0.198 & 0.376 &0.441& 0.243& 0.043 &0.441& 0.532\\ \hline
DistMult \cite{yang2014embedding} $\spadesuit$&0.241 & 0.155 & 0.263 & 0.419 & 0.430& 0.390 &0.440 &0.490\\ \hline
ComplEx \cite{trouillon2016complex} $\spadesuit$ &0.247 & 0.158 &0.275 &0.428 & 0.440 & 0.410 &0.460 &0.510\\ \hline
RotatE \cite{sun2019rotate} & 0.338& 0.241 &0.375& 0.533 & 0.476& 0.428 &0.492& 0.571\\ \hline
TuckER \citep{balavzevic2019tucker} & 0.358 &0.266 &0.394 &0.544 & 0.470 &0.443& 0.482& 0.526\\ \hline
HAKE \citep{zhang2020learning} & 0.346 &0.250& 0.381& 0.542&0.497& 0.452& 0.516& 0.582\\ \hline
CompGCN \citep{vashishth2019composition} & 0.355& 0.264 &0.390& 0.535& 0.479& 0.443& 0.494& 0.546\\ \hline
HittER\citep{chen-etal-2021-hitter}  & 0.344 &0.246 &0.380 &0.535 &0.496 &0.449 &0.514 &0.586\\ \hline

\multicolumn{9}{|c|}{\textbf{Text-based methods}}\\ \hline
Pretrain-KGE \citep{zhang2020pretrain}& 0.332&-& - &0.529 & 0.235 &-& -& 0.557 \\ \hline
KG-BERT \citep{yao2019kg} $\spadesuit$& - &-& - &0.420 & 0.216 &0.041 &0.302 &0.524 \\ \hline
StAR \citep{wang2021structure} $\spadesuit$ & 0.263 &0.171 &0.287& 0.452 &0.364 &0.222& 0.436& 0.647\\ \hline
MEM-KGC (w/o EP) \citep{choi2021mem} & 0.339 &0.249& 0.372& 0.522&  0.533 &0.473 &0.570 &0.636 \\ \hline
MEM-KGC (w/ EP) \citep{choi2021mem} & 0.346 &0.253& 0.381 &0.531 & \underline{0.557} & \underline{0.475} & \underline{0.604} & \textbf{0.704} \\ \hline

\multicolumn{9}{|c|}{\textbf{LLM-based methods}}\\ \hline
ChatGPT$_{zero-shot}$ \citep{zhu2023llms} $\diamond$& - & 0.237   &- & - & - & 0.190 &- &-\\ \hline
ChatGPT$_{one-shot}$ \citep{zhu2023llms} $\diamond$& - & 0.267   &- & - & - & 0.212 &- &-\\ \hline
KICGPT & \textbf{0.412}  & \textbf{0.327}   & \textbf{0.448} & \underline{0.554} & 0.549  & 0.474  & 0.585 &  0.641\\ \hline
KICGPT$_{tsa}$ & \underline{0.410}  & \underline{0.321} &    \underline{0.430}  &\textbf{0.581}&  	      \textbf{0.564}  & \textbf{0.478}  &   \textbf{0.612} &      \underline{0.677}\\ \hline

\end{tabular}}
\caption{ Comparison between the proposed methods and baseline methods. The best result in terms of each metric is shown in \textbf{bold} and the second best one is \underline{underlined}. $\spadesuit$ indicates that results are copied from \citep{wang2022language}, $\clubsuit$ implies that results are copied from \citep{chen-etal-2021-hitter}, $\diamond$ means that the results are running on the entire data according to the settings in \citep{zhu2023llms}, and other results are taken from their original papers. 
EP denotes the entity prediction task in the MEM-KGC model.}
\label{table:main}
\end{table*}

In this section, we empirically evaluate KICGPT.

\subsection{Setup}

\paragraph{Datasets.} We evaluate the proposed methods on the FB15k-237
\citep{toutanova2015representing} and WN18RR \citep{dettmers2018convolutional},
which are widely-used benchmark for link prediction. FB15k-237 is a subset of
the freebase \citep{bollacker2008freebase} knowledge graph, which includes commonsense
knowledge about topics such as movies, sports, awards, and traveling. WN18RR is a
subset of WordNet \citep{miller1995wordnet}, which contains knowledge about
English morphology. Both the FB15k-237 and WN18RR datasets remove redundant
inverse relations in case of information leakage. Compared with FB15k-237, 
the knowledge graph of
WN18RR
is much sparser.
The statistics of the two datasets are
shown in Table \ref{table:datasets}.

\begin{table}[!ht]
\resizebox{\columnwidth}{!}{
\begin{tabular}{cccccc}
\toprule 
Dataset & \# Ent  & \# Rel & \# train & \# valid & \# test \\ 
\midrule 
WN18RR & 40,943 & 11 & 86,835 & 3,034 & 3,134 \\
FB15k-237 & 14,541 & 237 & 272,115 & 17,535 & 20,466 \\
\bottomrule 
\end{tabular}
}
\caption{Statistics of the benchmark datasets.}
\label{table:datasets}
\end{table}
\paragraph{Baselines.} 
We compare the proposed KICGPT with a number of triple-based, text-based and LLM-based baselines.
The triple-based baselines include
RESCAL \citep{nickel2011three},
TransE \citep{bordes2013translating}, DistMult \citep{yang2014embedding}, ComplEx \citep{trouillon2016complex}, RotatE \citep{sun2019rotate}, TuckER \citep{balavzevic2019tucker}, HAKE \citep{zhang2020learning}, CompGCN \citep{vashishth2019composition}, and HittER \citep{chen-etal-2021-hitter}. The
text-based baselines\footnote{We did not compare our method with
\citeauthor{wang2022language} (\citeyear{wang2022language}) due to a data leakage
issue on the entity degrees.} include
Pretrain-KGE \citep{zhang2020pretrain}, KG-BERT \citep{yao2019kg}, StAR
\citep{wang2021structure}, and MEM-KGC
\citep{choi2021mem}. The LLM-based baselines are based on ChatGPT.
\citeauthor{zhu2023llms} (\citeyear{zhu2023llms}) reported zero-shot and one-shot link prediction results
on 25 instances sampled from FB15k-237. We run their codes on the whole
dataset and report the updated results as ChatGPT$_{zero-shot}$ and ChatGPT$_{one-shot}$. 

\paragraph{Implementation details.}  Though the proposed method can be used with various retrievers and
LLMs,
we prefer lightweight
models for efficiency considerations. Specifically,
our
KICGPT implementation uses RotatE \citep{sun2019rotate} as the retriever
and ChatGPT (gpt-3.5-turbo) as the LLM.
Hyper-parameters for RotatE 
are set as in \cite{sun2019rotate}. 
The value of $m$ 
is selected from
\{10, 20, 30, 40, 50\}, and the batch size from \{4, 8, 16, 32\} on a
small random subset with 200 instances from the validation set.
For the ChatGPT API, we set
temperature,
presence\_penalty, and
frequency\_penalty to 0, and top\_p to 1 to avoid randomness. 
The detailed prompts are shown in Appendix \ref{sec:appendix}.

\paragraph{Metrics.} Link prediction outputs a ranked list of all KG entities.
We report
Mean Reciprocal Rank (MRR), and Hits@{1, 3, 10} of the ranked
list under the ``filtered'' setting. The ``filtered'' setting
\citep{bordes2013translating} is a common practice that filters out
valid entities other than ground truth target entities from the ranked list.

\subsection{Experimental Results}

Results are shown in Table \ref{table:main}.\footnote{We only report
Hits@1 for the
LLM-based baselines because 
the LLM 
only outputs an answer
but not a list.
} 
As can be seen, the proposed KICGPT and KICGPT$_{tsa}$ achieve
state-of-the-art performance on both FB15k-237 and WN18RR in most
metrics.

Specifically, on the Fb15k-237 dataset, both variants of KICGPT surpass all the
baselines in terms of all the evaluation metrics. The text self-alignment method helps improve the 
Hits@10 
performance 
but degrades the 
MRR and Hits@\{1,3\}
performance. This may be because these demonstrations and aligned text are mutually supportive, which enhances the confidence of LLM to understand correct semantics, but for relations that LLM does not understand well, the aligned texts may not convey the true semantics.

On the Fb15k-237 dataset, both KICGPT and KICGPT$_{tsa}$ surpass all the
baselines in terms of all evaluation metrics. Text self-alignment 
helps improve the 
Hits@10 
performance 
but degrades the 
MRR and Hits@\{1,3\}
performance. This may be because the text self-alignment texts are derived from the demonstrations, which are also used in the LLM inference process along with the text self-alignment texts. While the mutually supportive text self-alignment texts and demonstrations enhance the LLM's understanding of correct semantics, they also make incorrect interpretations of relational semantics more persistent.

\begin{table}[!ht]
\resizebox{\columnwidth}{!}
{
\begin{tabular}{|c|c|}\hline
WN18RR & FB15k-237\\ \hline
\_also\_see & /tv/tv\_program/country\_of\_origin\\ \hline 
\_hypernym & /location/location/partially\_contains\\ \hline
\_has\_part & \makecell{/common/topic/webpage. \\/common/webpage/category}\\ \hline
\end{tabular}
}
\caption{Example relations from the WN18RR and FB15k-237 datasets.}
\label{table_relation_examples}
\end{table}

On the WN18RR dataset, KICGPT$_{tsa}$ achieves state-of-the-art performance 
on all metrics except Hits@10. 
Unlike the FB15k-237 dataset,
text self-alignment improves all metrics. 
This may be partly because the higher average number of triples per relation in
WN18RR (8454.8) compared to FB15k-237 (1308.5), which facilitates the generation of more precise text descriptions due to the increased availability of information for each relation.
Besides, as illustrated in Table \ref{table_relation_examples},  WN18RR relations
have more concise and direct semantics than those in FB15k-237, making text
self-alignment for WN18RR easier than for FB15k-237. These differences cause the
LLM to summarize more concise and precise descriptions for relations from WN18RR, which boosts the performance with higher-quality aligned text.  

The proposed methods show better performance over the triple-based 
and text-based baselines. This demonstrates the usefulness of the knowledge base
inside the LLM. Besides, compared with the LLM-based baselines 
(i.e., ChatGPT baselines\footnote{The results we report for ChatGPT baselines are
worse than those reported in \cite{zhu2023llms}. This is because they experimented
on only 25 triples on the FB15k-237 dataset while we test on the whole 20,466
testing triples. Besides, they manually restated descriptions in KG as clean and
natural expressions for the 25 triple, but doing this for all the testing triples
is labor-intensive and time-costly. In our experiment, we use the text generated
by text self-alignment as a substitution.}), the proposed methods
significantly outperform on both datasets, which shows the superiority of
integrating KG with LLMs for KGC. The performance improvement mainly comes from
the injection of information contained in the knowledge graph. 

The demonstration ordering process takes 50.13 and 30.29 minutes for the FB15k-237 and WN18RR datasets, respectively. In terms of training efficiency, KICGPT takes only 28.36 and 20.63 minutes to train on FB15k-237 and WN18RR, respectively, making it more efficient compared to existing text-based methods that can take hours to train. This is because KICGPT does not require fine-tuning of the LLM.

\subsection{Ablation Study}
In this experiment,
using the FB15k-237 dataset,
we perform ablation studies 
to demonstrate usefulness of each component in KICGPT.
Table \ref{table:ablation}
shows
the results.

In the first ablation study, we shuffle the top-$m$ entities in $R_{retriever}$ before
feeding into the LLM. As can be seen, this  causes a slight performance
degradation. This shows that the
order offered by the retriever is important and
can reduce the difficulty of re-ranking. Recently, \citeauthor{sun2023chatgpt}
(\citeyear{sun2023chatgpt}) also noted the significance of the initial order in search re-ranking.

In the
second study, we 
randomize the 
demonstration orders from the analogy and supplement pools.
Again, the performance degrades, demonstrating effectiveness of the proposed demonstration ordering in Section \ref{demon order}.

In the third study, we retrieve random triples from the KG as demonstrations. The
resultant performance degradation shows usefulness of the construction of demonstration pools in Section \ref{pool}. 

In the fourth study, we exclude all demonstrations, while still providing  the top-$m$
candidates to constrain the ChatGPT output. As can be seen, a significant
performance gap 
with the full model
is observed, 
demonstrating the necessity of the proposed ICL strategy. Note also that Hit@3
without ICL is superior to Hit@3 with random demonstrations. This may be
attributed to the vast number of triples in the KG, rendering  the
random demonstrations to have insignificant semantic relevance to the query. The
use of irrelevant triples as demonstrations to LLM can introduce noise,
potentially resulting in misleading outcomes.


In the last ablation study (Appendix \ref{sec:appendix}), we use a prompt without
the prompt engineering component (Section \ref{prompt}). This variant still uses
the proposed demonstration pools (Section \ref{pool}) and ordering (Section \ref{demon order}).
The observed degraded performance implies that the specially designed prompts can make use of the properties of the KGC task to boost performance.

\begin{table}[!ht]
\center
\resizebox{\linewidth}{!}{
\begin{tabular}{|l|c|c|c|c|}\hline

& \textbf{MRR} & \textbf{Hits@1} & \textbf{Hits@3} & \textbf{Hits@10} \\ \cline{1-5} \hline
KICGPT & \textbf{0.412} & \textbf{0.327} & \textbf{0.448} & \textbf{0.554} \\ \hline
\ \ shuffle candidates & 0.401  & 0.312   & 0.433 & 0.521\\ \hline
\ \  w/o demonstration ordering & 0.368 &	0.283 & 0.417 & 0.497 \\ \hline
\ \ random demonstrations & 0.349 &	0.271 & 0.387 &	0.481 \\ \hline
\ \  w/o ICL& 0.342 & 0.241 & 0.403 & 0.481\\ \hline
\ \  w/o prompt engineering & 0.401 & 0.307 & 0.432 & 0.548\\ \hline

\end{tabular}
}
\caption{Ablation results on
\textbf{FB15k-237}
(averaged over 4 random runs).}
\label{table:ablation}
\end{table}

\subsection{Analysis on Long-Tail Entities}

To show the effectiveness of KICGPT and KICGPT$_{tsa}$ to handle long-tail
entities, we follow \cite{wang2022language} and group entities by their
logarithm degrees in the knowledge graph.  Entities with lower degrees 
(and hence more likely to be long-tail entities)
are
assigned to groups with lower indexes.
A triple $(h, r, t)$ is considered relevant to group $d$ if $h$ or $t$ belongs to $d$. 

\begin{figure}[!ht]
    \includegraphics[width=\linewidth]{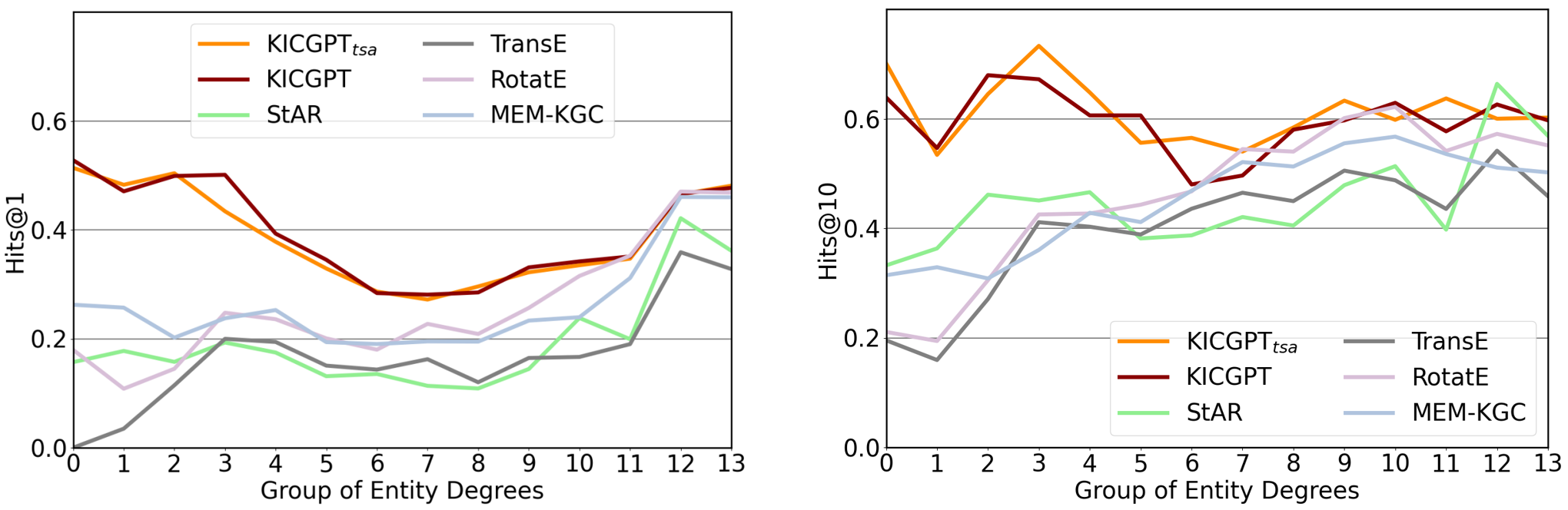}
    \caption{ Average performance of different models (in terms of Hits@1 and Hits@10) grouped by
the logarithm of entity degrees on the FB15k-237 dataset. }
    \label{fig: longtail}
\end{figure}

Figure \ref{fig: longtail} displays the Hits@1 and Hits@10 performance averages of various models on the FB15k-237 dataset, categorized by the logarithm of entity degrees. Text-based methods demonstrate slightly better performance than triple-based methods on long-tail entries. However, this improvement is not significant since it only applies to a small portion of long-tail entities (specifically, groups 0, 1, 2). 
Compared with these baselines, the proposed models achieve performance improvement
on almost all groups and perform significantly better on long-tail entities, 
which confirms the benefits of combining the LLM with KG.

\section{Conclusion}

In this paper, we propose KICGPT, an effective framework to integrate LLM and
traditional KGC methods for link prediction, and
a new ICL strategy called Knowledge Prompt. 
KICGPT utilizes the LLM as an extra
knowledge base.
Compared
with text-based methods, KICGPT utilizes the training-free property of
LLM to significantly reduce training overhead, and does not require finetuning
for different KGs. Experimental results show that KICGPT achieves
state-of-the-art performance on the link prediction benchmarks and is effective in
the handling of long-tail entities. 

\section*{Acknowledgements}

This work is supported by NSFC general grant 62076118 and Shenzhen fundamental research program JCYJ20210324105000003.

\section*{Limitations}

The performance improvement of the proposed approach in KGC is largely due to the
large and extensive knowledge base within the LLM. However, for some KGs (such as
personal preference data for users in an E-commerce platform), the LLM may not
contain enough relevant knowledge. Therefore, the proposed method works mostly for
commonsense KGs. Besides, because of the limited token length in the LLM, we
cannot inject all relevant facts from the KG as prompts. 

\section*{Ethics Statement}
There is no ethical problem in our study.
\bibliography{anthology,custom}

\begin{thebibliography}{45}
\expandafter\ifx\csname natexlab\endcsname\relax\def\natexlab#1{#1}\fi

\bibitem[{Bala{\v{z}}evi{\'c} et~al.(2019)Bala{\v{z}}evi{\'c}, Allen, and
  Hospedales}]{balavzevic2019tucker}
Ivana Bala{\v{z}}evi{\'c}, Carl Allen, and Timothy~M Hospedales. 2019.
\newblock Tucker: Tensor factorization for knowledge graph completion.
\newblock \emph{arXiv preprint arXiv:1901.09590}.

\bibitem[{Bang et~al.(2023)Bang, Cahyawijaya, Lee, Dai, Su, Wilie, Lovenia, Ji,
  Yu, Chung et~al.}]{bang2023multitask}
Yejin Bang, Samuel Cahyawijaya, Nayeon Lee, Wenliang Dai, Dan Su, Bryan Wilie,
  Holy Lovenia, Ziwei Ji, Tiezheng Yu, Willy Chung, et~al. 2023.
\newblock A multitask, multilingual, multimodal evaluation of chatgpt on
  reasoning, hallucination, and interactivity.
\newblock \emph{arXiv preprint arXiv:2302.04023}.

\bibitem[{Bollacker et~al.(2008)Bollacker, Evans, Paritosh, Sturge, and
  Taylor}]{bollacker2008freebase}
Kurt Bollacker, Colin Evans, Praveen Paritosh, Tim Sturge, and Jamie Taylor.
  2008.
\newblock Freebase: a collaboratively created graph database for structuring
  human knowledge.
\newblock In \emph{Proceedings of the 2008 ACM SIGMOD international conference
  on Management of data}, pages 1247--1250.

\bibitem[{Bordes et~al.(2013)Bordes, Usunier, Garcia-Duran, Weston, and
  Yakhnenko}]{bordes2013translating}
Antoine Bordes, Nicolas Usunier, Alberto Garcia-Duran, Jason Weston, and Oksana
  Yakhnenko. 2013.
\newblock Translating embeddings for modeling multi-relational data.
\newblock \emph{Advances in neural information processing systems}, 26.

\bibitem[{Brown et~al.(2020)Brown, Mann, Ryder, Subbiah, Kaplan, Dhariwal,
  Neelakantan, Shyam, Sastry, Askell, Agarwal, Herbert-Voss, Krueger, Henighan,
  Child, Ramesh, Ziegler, Wu, Winter, Hesse, Chen, Sigler, Litwin, Gray, Chess,
  Clark, Berner, McCandlish, Radford, Sutskever, and Amodei}]{gpt3}
Tom Brown, Benjamin Mann, Nick Ryder, Melanie Subbiah, Jared~D Kaplan, Prafulla
  Dhariwal, Arvind Neelakantan, Pranav Shyam, Girish Sastry, Amanda Askell,
  Sandhini Agarwal, Ariel Herbert-Voss, Gretchen Krueger, Tom Henighan, Rewon
  Child, Aditya Ramesh, Daniel Ziegler, Jeffrey Wu, Clemens Winter, Chris
  Hesse, Mark Chen, Eric Sigler, Mateusz Litwin, Scott Gray, Benjamin Chess,
  Jack Clark, Christopher Berner, Sam McCandlish, Alec Radford, Ilya Sutskever,
  and Dario Amodei. 2020.
\newblock \href
  {https://proceedings.neurips.cc/paper_files/paper/2020/file/1457c0d6bfcb4967418bfb8ac142f64a-Paper.pdf}
  {Language models are few-shot learners}.
\newblock In \emph{Advances in Neural Information Processing Systems},
  volume~33, pages 1877--1901. Curran Associates, Inc.

\bibitem[{Chen et~al.(2019)Chen, Zhang, Zhang, Chen, and Chen}]{chen2019meta}
Mingyang Chen, Wen Zhang, Wei Zhang, Qiang Chen, and Huajun Chen. 2019.
\newblock Meta relational learning for few-shot link prediction in knowledge
  graphs.
\newblock \emph{arXiv preprint arXiv:1909.01515}.

\bibitem[{Chen et~al.(2021)Chen, Liu, Gao, Jiao, Zhang, and
  Ji}]{chen-etal-2021-hitter}
Sanxing Chen, Xiaodong Liu, Jianfeng Gao, Jian Jiao, Ruofei Zhang, and Yangfeng
  Ji. 2021.
\newblock Hitter: Hierarchical transformers for knowledge graph embeddings.
\newblock In \emph{Proceedings of the 2021 Conference on Empirical Methods in
  Natural Language Processing (EMNLP)}. Association for Computational
  Linguistics.

\bibitem[{Choi et~al.(2021)Choi, Jang, and Ko}]{choi2021mem}
Bonggeun Choi, Daesik Jang, and Youngjoong Ko. 2021.
\newblock Mem-kgc: Masked entity model for knowledge graph completion with
  pre-trained language model.
\newblock \emph{IEEE Access}, 9:132025--132032.

\bibitem[{Dettmers et~al.(2018)Dettmers, Minervini, Stenetorp, and
  Riedel}]{dettmers2018convolutional}
Tim Dettmers, Pasquale Minervini, Pontus Stenetorp, and Sebastian Riedel. 2018.
\newblock Convolutional 2d knowledge graph embeddings.
\newblock In \emph{Proceedings of the AAAI conference on artificial
  intelligence}, volume~32.

\bibitem[{Dong et~al.(2022)Dong, Li, Dai, Zheng, Wu, Chang, Sun, Xu, and
  Sui}]{dong2022survey}
Qingxiu Dong, Lei Li, Damai Dai, Ce~Zheng, Zhiyong Wu, Baobao Chang, Xu~Sun,
  Jingjing Xu, and Zhifang Sui. 2022.
\newblock A survey for in-context learning.
\newblock \emph{arXiv preprint arXiv:2301.00234}.

\bibitem[{Jiang et~al.(2023)Jiang, Zhou, Dong, Ye, Zhao, and
  Wen}]{jiang2023structgpt}
Jinhao Jiang, Kun Zhou, Zican Dong, Keming Ye, Wayne~Xin Zhao, and Ji-Rong Wen.
  2023.
\newblock Structgpt: A general framework for large language model to reason
  over structured data.
\newblock \emph{arXiv preprint arXiv:2305.09645}.

\bibitem[{Liu et~al.(2021)Liu, Shen, Zhang, Dolan, Carin, and
  Chen}]{liu2021makes}
Jiachang Liu, Dinghan Shen, Yizhe Zhang, Bill Dolan, Lawrence Carin, and Weizhu
  Chen. 2021.
\newblock What makes good in-context examples for gpt-$3 $?
\newblock \emph{arXiv preprint arXiv:2101.06804}.

\bibitem[{Miller(1995)}]{miller1995wordnet}
George~A Miller. 1995.
\newblock Wordnet: a lexical database for english.
\newblock \emph{Communications of the ACM}, 38(11):39--41.

\bibitem[{Min et~al.(2022)Min, Lyu, Holtzman, Artetxe, Lewis, Hajishirzi, and
  Zettlemoyer}]{min2022rethinking}
Sewon Min, Xinxi Lyu, Ari Holtzman, Mikel Artetxe, Mike Lewis, Hannaneh
  Hajishirzi, and Luke Zettlemoyer. 2022.
\newblock Rethinking the role of demonstrations: What makes in-context learning
  work?
\newblock \emph{EMNLP}.

\bibitem[{Nickel et~al.(2011)Nickel, Tresp, Kriegel et~al.}]{nickel2011three}
Maximilian Nickel, Volker Tresp, Hans-Peter Kriegel, et~al. 2011.
\newblock A three-way model for collective learning on multi-relational data.
\newblock In \emph{Icml}, volume~11, pages 3104482--3104584.

\bibitem[{OpenAI(2023)}]{openai2023gpt}
OpenAI. 2023.
\newblock Gpt-4 technical report.
\newblock \emph{CoRR}.

\bibitem[{Ouyang et~al.(2022)Ouyang, Wu, Jiang, Almeida, Wainwright, Mishkin,
  Zhang, Agarwal, Slama, Ray et~al.}]{ouyang2022training}
Long Ouyang, Jeffrey Wu, Xu~Jiang, Diogo Almeida, Carroll Wainwright, Pamela
  Mishkin, Chong Zhang, Sandhini Agarwal, Katarina Slama, Alex Ray, et~al.
  2022.
\newblock Training language models to follow instructions with human feedback.
\newblock \emph{Advances in Neural Information Processing Systems},
  35:27730--27744.

\bibitem[{Peng et~al.(2023)Peng, Galley, He, Cheng, Xie, Hu, Huang, Liden, Yu,
  Chen et~al.}]{peng2023check}
Baolin Peng, Michel Galley, Pengcheng He, Hao Cheng, Yujia Xie, Yu~Hu, Qiuyuan
  Huang, Lars Liden, Zhou Yu, Weizhu Chen, et~al. 2023.
\newblock Check your facts and try again: Improving large language models with
  external knowledge and automated feedback.
\newblock \emph{arXiv preprint arXiv:2302.12813}.

\bibitem[{Rae et~al.(2021)Rae, Borgeaud, Cai, Millican, Hoffmann, Song,
  Aslanides, Henderson, Ring, Young et~al.}]{gopher}
Jack~W Rae, Sebastian Borgeaud, Trevor Cai, Katie Millican, Jordan Hoffmann,
  Francis Song, John Aslanides, Sarah Henderson, Roman Ring, Susannah Young,
  et~al. 2021.
\newblock Scaling language models: Methods, analysis \& insights from training
  gopher.
\newblock \emph{arXiv preprint arXiv:2112.11446}.

\bibitem[{Robertson and Walker(1994)}]{robertson1994some}
Stephen~E Robertson and Steve Walker. 1994.
\newblock Some simple effective approximations to the 2-poisson model for
  probabilistic weighted retrieval.
\newblock In \emph{SIGIR’94: Proceedings of the Seventeenth Annual
  International ACM-SIGIR Conference on Research and Development in Information
  Retrieval, organised by Dublin City University}, pages 232--241. Springer.

\bibitem[{Sadeghian et~al.(2019)Sadeghian, Armandpour, Ding, and
  Wang}]{sadeghian2019drum}
Ali Sadeghian, Mohammadreza Armandpour, Patrick Ding, and Daisy~Zhe Wang. 2019.
\newblock Drum: End-to-end differentiable rule mining on knowledge graphs.
\newblock \emph{Advances in Neural Information Processing Systems}, 32.

\bibitem[{Schick et~al.(2023)Schick, Dwivedi-Yu, Dess{\`\i}, Raileanu, Lomeli,
  Zettlemoyer, Cancedda, and Scialom}]{schick2023toolformer}
Timo Schick, Jane Dwivedi-Yu, Roberto Dess{\`\i}, Roberta Raileanu, Maria
  Lomeli, Luke Zettlemoyer, Nicola Cancedda, and Thomas Scialom. 2023.
\newblock Toolformer: Language models can teach themselves to use tools.
\newblock \emph{arXiv preprint arXiv:2302.04761}.

\bibitem[{Schlichtkrull et~al.(2018)Schlichtkrull, Kipf, Bloem, Van Den~Berg,
  Titov, and Welling}]{schlichtkrull2018modeling}
Michael Schlichtkrull, Thomas~N Kipf, Peter Bloem, Rianne Van Den~Berg, Ivan
  Titov, and Max Welling. 2018.
\newblock Modeling relational data with graph convolutional networks.
\newblock In \emph{The Semantic Web: 15th International Conference, ESWC 2018,
  Heraklion, Crete, Greece, June 3--7, 2018, Proceedings 15}, pages 593--607.
  Springer.

\bibitem[{Sun et~al.(2023)Sun, Yan, Ma, Ren, Yin, and Ren}]{sun2023chatgpt}
Weiwei Sun, Lingyong Yan, Xinyu Ma, Pengjie Ren, Dawei Yin, and Zhaochun Ren.
  2023.
\newblock Is chatgpt good at search? investigating large language models as
  re-ranking agent.
\newblock \emph{arXiv preprint arXiv:2304.09542}.

\bibitem[{Sun et~al.(2019)Sun, Deng, Nie, and Tang}]{sun2019rotate}
Zhiqing Sun, Zhi-Hong Deng, Jian-Yun Nie, and Jian Tang. 2019.
\newblock Rotate: Knowledge graph embedding by relational rotation in complex
  space.
\newblock \emph{ICLR}.

\bibitem[{Tay et~al.(2022)Tay, Wei, Chung, Tran, So, Shakeri, Garcia, Zheng,
  Rao, Chowdhery et~al.}]{tay2022transcending}
Yi~Tay, Jason Wei, Hyung~Won Chung, Vinh~Q Tran, David~R So, Siamak Shakeri,
  Xavier Garcia, Huaixiu~Steven Zheng, Jinfeng Rao, Aakanksha Chowdhery, et~al.
  2022.
\newblock Transcending scaling laws with 0.1\% extra compute.
\newblock \emph{arXiv preprint arXiv:2210.11399}.

\bibitem[{Toutanova et~al.(2015)Toutanova, Chen, Pantel, Poon, Choudhury, and
  Gamon}]{toutanova2015representing}
Kristina Toutanova, Danqi Chen, Patrick Pantel, Hoifung Poon, Pallavi
  Choudhury, and Michael Gamon. 2015.
\newblock Representing text for joint embedding of text and knowledge bases.
\newblock In \emph{Proceedings of the 2015 conference on empirical methods in
  natural language processing}, pages 1499--1509.

\bibitem[{Trouillon et~al.(2016)Trouillon, Welbl, Riedel, Gaussier, and
  Bouchard}]{trouillon2016complex}
Th{\'e}o Trouillon, Johannes Welbl, Sebastian Riedel, {\'E}ric Gaussier, and
  Guillaume Bouchard. 2016.
\newblock Complex embeddings for simple link prediction.
\newblock In \emph{International conference on machine learning}, pages
  2071--2080. PMLR.

\bibitem[{Vashishth et~al.(2019)Vashishth, Sanyal, Nitin, and
  Talukdar}]{vashishth2019composition}
Shikhar Vashishth, Soumya Sanyal, Vikram Nitin, and Partha Talukdar. 2019.
\newblock Composition-based multi-relational graph convolutional networks.
\newblock \emph{ICLR}.

\bibitem[{Wang et~al.(2021{\natexlab{a}})Wang, Shen, Long, Zhou, Wang, and
  Chang}]{wang2021structure}
Bo~Wang, Tao Shen, Guodong Long, Tianyi Zhou, Ying Wang, and Yi~Chang.
  2021{\natexlab{a}}.
\newblock Structure-augmented text representation learning for efficient
  knowledge graph completion.
\newblock In \emph{Proceedings of the Web Conference 2021}, pages 1737--1748.

\bibitem[{Wang et~al.(2017)Wang, Mao, Wang, and Guo}]{wang2017knowledge}
Quan Wang, Zhendong Mao, Bin Wang, and Li~Guo. 2017.
\newblock Knowledge graph embedding: A survey of approaches and applications.
\newblock \emph{IEEE Transactions on Knowledge and Data Engineering},
  29(12):2724--2743.

\bibitem[{Wang et~al.(2021{\natexlab{b}})Wang, Gao, Zhu, Zhang, Liu, Li, and
  Tang}]{wang2021kepler}
Xiaozhi Wang, Tianyu Gao, Zhaocheng Zhu, Zhengyan Zhang, Zhiyuan Liu, Juanzi
  Li, and Jian Tang. 2021{\natexlab{b}}.
\newblock Kepler: A unified model for knowledge embedding and pre-trained
  language representation.
\newblock \emph{Transactions of the Association for Computational Linguistics},
  9:176--194.

\bibitem[{Wang et~al.(2022{\natexlab{a}})Wang, He, Liang, and
  Xiao}]{wang2022language}
Xintao Wang, Qianyu He, Jiaqing Liang, and Yanghua Xiao. 2022{\natexlab{a}}.
\newblock Language models as knowledge embeddings.
\newblock \emph{IJCAI}.

\bibitem[{Wang et~al.(2022{\natexlab{b}})Wang, Wei, Schuurmans, Le, Chi, and
  Zhou}]{wang2022rationale}
Xuezhi Wang, Jason Wei, Dale Schuurmans, Quoc Le, Ed~Chi, and Denny Zhou.
  2022{\natexlab{b}}.
\newblock Rationale-augmented ensembles in language models.
\newblock \emph{arXiv preprint arXiv:2207.00747}.

\bibitem[{Wei et~al.(2022)Wei, Tay, Bommasani, Raffel, Zoph, Borgeaud,
  Yogatama, Bosma, Zhou, Metzler, Chi, Hashimoto, Vinyals, Liang, Dean, and
  Fedus}]{wei2022emergent}
Jason Wei, Yi~Tay, Rishi Bommasani, Colin Raffel, Barret Zoph, Sebastian
  Borgeaud, Dani Yogatama, Maarten Bosma, Denny Zhou, Donald Metzler, Ed~H.
  Chi, Tatsunori Hashimoto, Oriol Vinyals, Percy Liang, Jeff Dean, and William
  Fedus. 2022.
\newblock \href {https://openreview.net/forum?id=yzkSU5zdwD} {Emergent
  abilities of large language models}.
\newblock \emph{Transactions on Machine Learning Research}.
\newblock Survey Certification.

\bibitem[{Xie et~al.(2016)Xie, Liu, Jia, Luan, and Sun}]{xie2016representation}
Ruobing Xie, Zhiyuan Liu, Jia Jia, Huanbo Luan, and Maosong Sun. 2016.
\newblock Representation learning of knowledge graphs with entity descriptions.
\newblock In \emph{Proceedings of the AAAI Conference on Artificial
  Intelligence}, volume~30.

\bibitem[{Xiong et~al.(2018)Xiong, Yu, Chang, Guo, and Wang}]{xiong2018one}
Wenhan Xiong, Mo~Yu, Shiyu Chang, Xiaoxiao Guo, and William~Yang Wang. 2018.
\newblock One-shot relational learning for knowledge graphs.
\newblock \emph{arXiv preprint arXiv:1808.09040}.

\bibitem[{Yang et~al.(2014)Yang, Yih, He, Gao, and Deng}]{yang2014embedding}
Bishan Yang, Wen-tau Yih, Xiaodong He, Jianfeng Gao, and Li~Deng. 2014.
\newblock Embedding entities and relations for learning and inference in
  knowledge bases.
\newblock \emph{arXiv preprint arXiv:1412.6575}.

\bibitem[{Yao et~al.(2019)Yao, Mao, and Luo}]{yao2019kg}
Liang Yao, Chengsheng Mao, and Yuan Luo. 2019.
\newblock Kg-bert: Bert for knowledge graph completion.
\newblock \emph{arXiv preprint arXiv:1909.03193}.

\bibitem[{Ye et~al.(2023)Ye, Wu, Feng, Yu, and Kong}]{ye2023compositional}
Jiacheng Ye, Zhiyong Wu, Jiangtao Feng, Tao Yu, and Lingpeng Kong. 2023.
\newblock Compositional exemplars for in-context learning.
\newblock \emph{arXiv preprint arXiv:2302.05698}.

\bibitem[{Yu et~al.(2018)Yu, Zhang, Yang, Yasunaga, Wang, Li, Ma, Li, Yao,
  Roman et~al.}]{yu2018spider}
Tao Yu, Rui Zhang, Kai Yang, Michihiro Yasunaga, Dongxu Wang, Zifan Li, James
  Ma, Irene Li, Qingning Yao, Shanelle Roman, et~al. 2018.
\newblock Spider: A large-scale human-labeled dataset for complex and
  cross-domain semantic parsing and text-to-sql task.
\newblock \emph{Proceedings of the 2018 Conference on Empirical Methods in
  Natural Language Processing}.

\bibitem[{Zhang et~al.(2020{\natexlab{a}})Zhang, Cai, Zhang, and
  Wang}]{zhang2020learning}
Zhanqiu Zhang, Jianyu Cai, Yongdong Zhang, and Jie Wang. 2020{\natexlab{a}}.
\newblock Learning hierarchy-aware knowledge graph embeddings for link
  prediction.
\newblock In \emph{Proceedings of the AAAI conference on artificial
  intelligence}, volume~34, pages 3065--3072.

\bibitem[{Zhang et~al.(2020{\natexlab{b}})Zhang, Liu, Zhang, Su, Sun, and
  He}]{zhang2020pretrain}
Zhiyuan Zhang, Xiaoqian Liu, Yi~Zhang, Qi~Su, Xu~Sun, and Bin He.
  2020{\natexlab{b}}.
\newblock Pretrain-kge: learning knowledge representation from pretrained
  language models.
\newblock In \emph{Findings of the Association for Computational Linguistics:
  EMNLP 2020}, pages 259--266.

\bibitem[{Zhao et~al.(2021)Zhao, Wallace, Feng, Klein, and
  Singh}]{zhao2021calibrate}
Zihao Zhao, Eric Wallace, Shi Feng, Dan Klein, and Sameer Singh. 2021.
\newblock Calibrate before use: Improving few-shot performance of language
  models.
\newblock In \emph{International Conference on Machine Learning}, pages
  12697--12706. PMLR.

\bibitem[{Zhu et~al.(2023)Zhu, Wang, Chen, Qiao, Ou, Yao, Deng, Chen, and
  Zhang}]{zhu2023llms}
Yuqi Zhu, Xiaohan Wang, Jing Chen, Shuofei Qiao, Yixin Ou, Yunzhi Yao, Shumin
  Deng, Huajun Chen, and Ningyu Zhang. 2023.
\newblock Llms for knowledge graph construction and reasoning: Recent
  capabilities and future opportunities.
\newblock \emph{arXiv preprint arXiv:2305.13168}.

\end{thebibliography}
\bibliographystyle{acl_natbib}
\newcommand{\code}[1]{\colorbox[RGB]{245,245,245}{\texttt{\detokenize{#1}}}}

\clearpage

\appendix
\section{Appendix}
\subsection{KICGPT Algorithm}
\label{app: alg}
We provide the algorithm of KICGPT here.

\begin{algorithm*}[hb]
\DontPrintSemicolon
\newcommand\mycommfont[1]{\footnotesize\ttfamily{#1}}
\SetCommentSty{mycommfont}

\SetKwInput{KwInput}{Input}                
\SetKwInput{KwOutput}{Output}              
 
  \KwInput{KG $G = {(h,r,t)}$, $E$, and $R$ denote the entity set and the relation set of $G$, $G_{train}$ and $G_{valid}$ are sets of triples in $G$ for training and validation\newline link prediction query $q = (h,r,?)$}
  \KwOutput{analogy pool $D_a$\newline supplement pool $D_s$}
  $D_{a} = \{ (e',r,e'') \in G_{train} \cup G_{valid}| e',e''\in E\}$\\
  $D_{s} = \{(h,r',e') \in G_{train} \cup G_{valid}|r'\in R, e'\in E\} \cup \{(e',r',h) \in G_{train} \cup G_{valid}|r'\in R, e'\in E\}$

\caption{Demonstration pools from KG}
\end{algorithm*}

\SetKwRepeat{Do}{do}{while}

\begin{algorithm*}[hb]
\DontPrintSemicolon
\newcommand\mycommfont[1]{\footnotesize\ttfamily{#1}}
\SetCommentSty{mycommfont}

\SetKwInput{KwInput}{Input}                
\SetKwInput{KwOutput}{Output}              
 
  \KwInput{Analogy demonstration pool: $D_{a}$, entity set $E$ of KG }
  \KwOutput{ordered list $L_{a}$}
    \ForEach{element $e \in E$}{initialize a zero counter $c_e=0$ for $e$}
    Randomly select a triple $(h',r',t')$ from $D_a$, append it to $L_a$ and increase the counter $c_{h'}$ and $c_{t'}$ by 1\\
    \Repeat{$D_{a}$ is empty}{
        Find a triple $a$ that has the minimum sum of $c_{e'}$ and $c_{e''}$ where $e'$ and $e''$ are the head and tail entity in triple $a$ (If there are multiple triples with minimum sum of entity counters, randomly select one of them as $a$)\\
        Append $a$ to $L_a$ and remove $a$ from $D_{a}$\\
        increase the counter of entities $c_{e'}, c_{e''}$ in triple $a$ by 1
    }
\caption{Diversity-based ordering for analogy pool}
\end{algorithm*}

\begin{algorithm*}[hb]
\DontPrintSemicolon
\newcommand\mycommfont[1]{\footnotesize\ttfamily{#1}}
\SetCommentSty{mycommfont}

\SetKwInput{KwInput}{Input}                
\SetKwInput{KwOutput}{Output}              
 
  \KwInput{Supplement demonstration pool: $D_{s}$ \newline link prediction query $q$}
  \KwOutput{ordered list $L_{s}$}
  \lForEach{triple $s$ in $D_{s}$}{\newline calculate the BM25 score between text descriptions of $s$ and $q$}
  $L_s\leftarrow$Ordering triples in $D_s$ according to the BM25 score in descending order
\caption{BM25 score-based ordering for supplement pool}
\end{algorithm*}

\begin{algorithm*}[hb]
\DontPrintSemicolon
\newcommand\mycommfont[1]{\footnotesize\ttfamily{#1}}
\SetCommentSty{mycommfont}

\SetKwInput{KwInput}{Input}                
\SetKwInput{KwOutput}{Output}              
   \tcc{For presentation simplicity, we represent the algorithm on missing tail queries. missing head queries are handled in a similar way.}
  \KwInput{KG $G = {(h,r,t)}$, $E$, and $R$ denote the entity set and the relation set of $G$, $G_{train}$ and $G_{valid}$ are sets of triples in $G$ for training and validation\newline link prediction query $q = (h,r,?)$}
  \KwOutput{Ranked list $R_{KICGPT}$ contains all $e \in E$}

  Ranked list $R_{retriever}$ contains all $e \in E$ $\leftarrow$ Handling $q$ by triple-based KGC retriever\newline $R_{retriever} = [e_{1st},e_{2nd},e_{3rd},\dots, e_{|E|th}]$\\

  Analogy Pool $D_a$, Supplement Pool $D_s$ $\leftarrow$ Fetch Demonstration Pools from G based on $q = (h,r,?)$  
    \\
  $L_a\leftarrow$ Ordering triples in $D_a$ based on diversity \\
  $L_s\leftarrow$ Ordering triples in $D_s$ based on BM25 score\\

  Use a unified prompt template to convert the query $q$ and triples in $L_a$ and $L_s$ to text\\

  With some instructions, demonstrations arranged by order in $L_a$ and $L_s$ are organized to ICL prompts\\

  Given the ICL prompts and query, LLM is asked to perform re-ranking for first m entities in $R_{retriever}$  \newline
  $R_{LLM}\leftarrow$ LLM (first m entities in $R_{retriever}$, ICL prompts, text of $q$)
  \newline $R_{LLM} = [e'_{1st},e'_{2nd},e'_{3rd},\dots,e'_{mth}]$, is permutation of first m entities in $R_{retriever}$
  \\
$R_{KICGPT} = [e'_{1st},e'_{2nd},e'_{3rd},\dots,e'_{mth},e_{m+1th},\dots,e_{|E|th}]$, where first m entities is $R_{LLM}$ and remaining come from corresponding location of $R_{retriever}$
\caption{Algorithm for KICGPT}
\end{algorithm*}

\clearpage
\subsection{Prompt Formatting}
\label{sec:appendix}

We list the prompts used in this paper as follows.

\begin{table*}[hbp]
   
  \footnotesize 
  \setlength\LTleft{-30pt}%
  \setlength\LTright{-30pt}%
  \centering 
  	\setlength\tabcolsep{1.5pt}
	\renewcommand{\arraystretch}{2}
\resizebox{\linewidth}{!}{
  \begin{tabular}
  {p{1.5in}|p{4in}|p{1.5in}} \toprule
 \textbf{Datasets}

		&  \textbf{Prompt Template} & \textbf{ChatGPT}   \\
		\midrule

FB15k-237 & \texttt{\detokenize{You are a good assistant to reading, understanding and summarizing. }}\newline
\texttt{\detokenize{Ecuador is the partially_contains of location of location of Pacific Ocean}}\newline
\texttt{\detokenize{Appalachian Mountains is the partially_contains of location of location of Massachusetts}}\newline
\texttt{\detokenize{Moldavia is the partially_contains of location of location of Moldova}}\newline
\texttt{\detokenize{Ecuador is the partially_contains of location of location of South America}}\newline
\texttt{\detokenize{Adirondack Mountains is the partially_contains of location of location of Warren County}}\newline
\texttt{\detokenize{In above examples, What do you think "partially_contains of location of location of " mean? Summarize and descript its meaning using the format: "If the example shows something A is partially_contains of location of location of  of something B, it means A is [mask] of B." Fill the mask and the statement should be as short as possible.}}

& If the example shows something A is partially\_contains of location of location of of something B, it means A is located partially within the boundaries of B.
  \\ \hline
  WN18RR & \texttt{\detokenize{You are a good assistant to reading, understanding and summarizing.}} \newline
\texttt{\detokenize{red indian be member of domain usage of disparagement
compass be member of domain usage of archaism}} \newline
\texttt{\detokenize{penicillin v potassium be member of domain usage of trade name}} \newline
\texttt{\detokenize{nanna be member of domain usage of united kingdom of great britain and northern ireland }}\newline
\texttt{\detokenize{nerves be member of domain usage of plural form}} \newline
\texttt{\detokenize{zyloprim be member of domain usage of trademark}} \newline
\texttt{\detokenize{
In above examples, What do you think "be member of domain usage of" mean? Summarize and descript its meaning using the format: "If the example shows something A be member of domain usage of something B, it means A is [mask] of B." Fill the mask and the statement should be as short as possible.}}

  & 
  If the example shows something A be member of domain usage of something B, it means A is a term or word that belongs to the category or domain of B's usage.
  \\ \hline
  \end{tabular}

  }
    \caption{Examples of prompts for text self-alignment.}
    \label{table:text_align_prompt}
  \end{table*}

\begin{table*}[hbp]
   
  \footnotesize 
  \setlength\LTleft{-30pt}%
  \setlength\LTright{-30pt}%
  \centering 
  	\setlength\tabcolsep{1.5pt}
	\renewcommand{\arraystretch}{2}
\resizebox{\linewidth}{!}{
  \begin{tabular}
  {p{1.5in}|p{4in}|p{1.5in}} \toprule
 \textbf{Prompt Formatting}

		&  \textbf{Prompt Template} & \textbf{ChatGPT}   \\
		\midrule

Trivial Prompt & \texttt{\detokenize{predict the tail entity [MASK] from the given (Mel Blanc,type_of_union of marriage of people of spouse_s  of person of people of , [MASK]) by completing the sentence "what is the type_of_union of marriage of people of spouse_s  of person of people of Mel Blanc? The answer is ". The answer is Marriage, so the [MASK] is Marriage.
predict the tail entity [MASK] from the given (Screen Actors Guild Life Achievement Award,award_winner of award_honor of award of winners  of award_category of award of , [MASK]) by completing the sentence "what is the award_winner of award_honor of award of winners  of award_category of award of Screen Actors Guild Life Achievement Award? The answer is ". The answer is Stan Laurel, so the [MASK] is Stan Laurel.}} \newline
\texttt{\detokenize{
The list of candidate answers is [Marriage, Domestic partnership, Civil union, Official Website, Rang De Basanti, HBO, Male, Television, Judaism-GB, Crusades]. And the question is predict the tail entity [MASK] from the given (Stan Laurel,type_of_union of marriage of people of spouse_s  of person of people of , [MASK]) by completing the sentence "what is the type_of_union of marriage of people of spouse_s  of person of people of Stan Laurel? The answer is ". Now, based on the previous examples and your own knowledge and thinking, sort the list to let the candidate answers which are more possible to be the true answer to the question prior. Output the sorted order of candidate answers using the format "[most possible answer | second possible answer | ... | least possible answer]" and please start your response with "The final order:". Do not output anything except the final order. Note your output sorted order should contain all the candidates in the list but not add new answers to it. }}

& The final order: [Marriage | Domestic partnership | Civil union | Official Website | HBO | Male | Television | Rang De Basanti | Judaism-GB | Crusades].
  \\ \hline
  \end{tabular}}
    \caption{Trivial Prompts on the FB15k-237 dataset.}
    \label{table:trivial_prompt}
  \end{table*}

\begin{table*}[hb]
   
  \footnotesize 
  \setlength\LTleft{-30pt}%
  \setlength\LTright{-30pt}%
  \centering 
  	\setlength\tabcolsep{1.5pt}
	\renewcommand{\arraystretch}{2}
\resizebox{\linewidth}{!}{
  \begin{tabular}
  {p{1.5in}|p{4in}|p{1.5in}} \toprule
 \textbf{Steps}

		&  \textbf{Prompt Template} & \textbf{ChatGPT}   \\
		\midrule
		Responsibility Description
		& \texttt{\detokenize{You are a good assistant to perform link prediction and sorting. Given a goal question and a list of candidate answers to this question. You need to order these candidate answers in the list to let candidate answers which are more possible to be the answer to the question prior. If you have known your responsibility, respond "Yes". Otherwise, respond "No". Do not output anything except "Yes" and "No".}} 
		& Yes.
	\\ \cline{1-3}
        Question and Demonstration Description & \texttt{\detokenize{The goal question is: predict the tail entity [MASK] from the given (Stan Laurel,type_of_union of marriage of people of spouse_s  of person of people of, [MASK]) by completing the sentence "what is the type_of_union of the marriage of people of spouse_s  of the person of people of Stan Laurel? The answer is ". To sort the candidate answers, typically you would need to refer to some other examples that may be similar to or related to the question.
 Part of the given examples are similar to the goal question, you should analogy them to understand the potential meaning of the goal question. Another part of the given facts contains supplementary information, keep capturing this extra information and mining potential relationships among them to help the sorting. Please carefully read, realize, and think about these examples. Summarize the way of thinking in these examples and memorize the information you think maybe help your sorting task. During I give examples please keep silent until I let you output.}}
& Okay, I understand. I will wait for your examples and instructions.  \\ \cline{1-3}
Multiple Demonstrations & \texttt{\detokenize{Examples used to Analogy: "predict the tail entity [MASK] from the given (Mel Blanc,type_of_union of marriage of people of spouse_s  of person of people of , [MASK]) by completing the sentence "what is the type_of_union of marriage of people of spouse_s  of person of people of Mel Blanc? The answer is ". The answer is Marriage, so the [MASK] is Marriage."}} \newline \newline
\texttt{\detokenize{Examples give supplement information: "predict the tail entity [MASK] from the given (Screen Actors Guild Life Achievement Award,award_winner of award_honor of award of winners  of award_category of award of , [MASK]) by completing the sentence "what is the award_winner of award_honor of award of winners  of award_category of award of Screen Actors Guild Life Achievement Award? The answer is ". The answer is Stan Laurel, so the [MASK] is Stan Laurel.  Keep thinking but not output.}}
& Okay, I will keep thinking and analyzing the given examples to identify potential relationships and patterns that can help with the sorting task.
 \\ \cline{1-3} \hline
Final Query & \texttt{\detokenize{The list of candidate answers is [Marriage, Domestic partnership, Civil union, Official Website, Rang De Basanti, HBO, Male, Television, Judaism-GB, Crusades]. And the question is predict the tail entity [MASK] from the given (Stan Laurel,type_of_union of marriage of people of spouse_s  of person of people of , [MASK]) by completing the sentence "what is the type_of_union of marriage of people of spouse_s  of person of people of Stan Laurel? The answer is ". Now, based on the previous examples and your own knowledge and thinking, sort the list to let the candidate answers which are more possible to be the true answer to the question prior. Output the sorted order of candidate answers using the format "[most possible answer | second possible answer | ... | least possible answer]" and please start your response with "The final order:". Do not output anything except the final order. Note your output sorted order should contain all the candidates in the list but not add new answers to it. }}

& The final order: [Marriage | Domestic partnership | Civil union | Official Website | HBO | Male | Television | Rang De Basanti | Judaism-GB | Crusades].
  \\ \hline
  \end{tabular}}
    \caption{Examples of prompts for KICGPT on the FB15k-237 dataset.}
    \label{table:fb15k-237_prompt}
  \end{table*}

\begin{table*}[hb]
   
  \footnotesize 
  \setlength\LTleft{-30pt}%
  \setlength\LTright{-30pt}%
  \centering 
  	\setlength\tabcolsep{1.5pt}
	\renewcommand{\arraystretch}{2}
\resizebox{\linewidth}{!}{
  \begin{tabular}
  {p{1.5in}|p{4in}|p{1.5in}}

    		\toprule \textbf{Steps}
		&  \textbf{Prompt Template} & \textbf{ChatGPT}   \\
		\midrule
		Responsibility Description
		& \texttt{\detokenize{Assume you're a linguist of English lexicons. You will be first given some examples. Then use these examples as references and your own knowledge to score for some statements. If you have known your responsibility, respond "Yes". Otherwise, respond "No". Do not output anything except "Yes" and "No".
}} 
		& Yes.
	\\ \cline{1-3}
        Question and Demonstration Description & \texttt{\detokenize{The goal statements are about member of domain usage of trade name. trade name : a name given to a product or service. Part of the given examples are similar to the statements, you should analogy them to understand the potential meaning of the statements to be scored. Another part of the given examples contains supplementary information, keep capturing this extra information and mining potential relationships among them to help the scoring. Please carefully read, realize and think about these examples. Summarize the way of thinking in these examples and memorize the information you think maybe help. DO NOT give me any feedback. }}
& Okay.  \\ \cline{1-3}
Multiple Demonstrations & \texttt{\detokenize{Examples used to Analogy:
trade name : a name given to a product or service. vinblastine : periwinkle plant derivative used as an antineoplastic drug (trade name Velban) that disrupts cell division. vinblastine be member of domain usage of trade name .
vernacular : a characteristic language of a particular group (as among thieves); "they don't speak our lingo". chink : a narrow opening as e.g. between planks in a wall. chink be member of domain usage of vernacular.}} \newline 
\texttt{\detokenize{Examples give supplement information:
trade name : a name given to a product or service. cortone acetate : a corticosteroid hormone (trade name Cortone Acetate) normally produced by the adrenal cortex; is converted to hydrocortisone. cortone acetate be member of domain usage of trade name .
trade name : a name given to a product or service. phenelzine : monoamine oxidase inhibitor (trade name Nardil) used to treat clinical depression. phenelzine be member of domain usage of trade name.
Keep thinking but DO NOT give me any feedback.
}} & Okay.

 \\ \cline{1-3} \hline
Final Query & \texttt{\detokenize{trade name : a name given to a product or service. verapamil : a drug (trade names Calan and Isoptin) used as an oral or parenteral calcium blocker in cases of hypertension or congestive heart failure or angina or migraine. verapamil be member of domain usage of trade name. Directly give a score out of 100 for the statement and DO NOT output any other thing}} \newline
... \newline
\texttt{\detokenize{trade name : a name given to a product or service. nitrostat : trade names for nitroglycerin used as a coronary vasodilator in the treatment of angina pectoris. nitrostat be member of domain usage of trade name .. Directly give a score out of 100 for the statement and DO NOT output any other thing}} \newline ... \newline \texttt{\detokenize{trade name : a name given to a product or service. hydantoin : any of a group of anticonvulsant drugs used in treating epilepsy. hydantoin be member of domain usage of trade name .. Directly give a score out of 100 for the statement and DO NOT output any other thing.}} \newline ...

& 90. \newline ... \newline 100\newline ... \newline 50 \newline ... 
  \\ \hline
  \end{tabular}}
    \caption{Examples of prompts  for KICGPT on the WN18RR dataset.}
    \label{table:wn18rr_prompt}
  \end{table*}

\end{document}